\documentclass[10pt,twocolumn,letterpaper]{article}

\usepackage{wacv}              

\usepackage[accsupp]{axessibility}
\usepackage{graphicx}
\usepackage{amsmath}
\usepackage{amssymb}
\usepackage{booktabs}
\usepackage{bm}
\usepackage{placeins} 

\usepackage{algorithm}
\usepackage{algpseudocode}
\MakeRobust{\Call}

\usepackage[pagebackref,breaklinks,colorlinks]{hyperref}

\usepackage[capitalize]{cleveref}
\crefname{section}{Sec.}{Secs.}
\Crefname{section}{Section}{Sections}
\Crefname{table}{Table}{Tables}
\crefname{table}{Tab.}{Tabs.}

\def\wacvPaperID{1303} 
\def\confName{WACV}
\def\confYear{2024}

\usepackage{tikz}
\usepackage{textcomp}
\usepackage{hyperref}
\usepackage{lipsum}

\newcommand\copyrighttext{%
  \footnotesize \textcopyright 2023 IEEE. Personal use of this material is permitted.
  Permission from IEEE must be obtained for all other uses, in any current or future 
  media, including reprinting/republishing this material for advertising or promotional 
  purposes, creating new collective works, for resale or redistribution to servers or 
  lists, or reuse of any copyrighted component of this work in other works.}
\newcommand\copyrightnotice{%
\begin{tikzpicture}[remember picture,overlay]
\node[anchor=south,yshift=10pt] at (current page.south) {\fbox{\parbox{\dimexpr\textwidth-\fboxsep-\fboxrule\relax}{\copyrighttext}}};
\end{tikzpicture}%
}

\begin{document}

\title{From Chaos to Calibration: A Geometric Mutual Information Approach to Target-Free Camera LiDAR Extrinsic Calibration}

\author{Jack Borer
\qquad
Jeremy Tschirner
\qquad
Florian Ölsner
\qquad
Stefan Milz
\\ Spleenlab GmbH
}
\maketitle

\copyrightnotice

\begin{abstract}
    Sensor fusion is vital for the safe and robust operation of autonomous vehicles. Accurate extrinsic sensor to sensor calibration is necessary to accurately fuse multiple sensor's data in a common spatial reference frame. In this paper, we propose a target free extrinsic calibration algorithm that requires no ground truth training data, artificially constrained motion trajectories, hand engineered features or offline optimization and that is accurate, precise and extremely robust to initialization error. 

    Most current research on online camera-LiDAR extrinsic calibration requires ground truth training data which is impossible to capture at scale. We revisit analytical mutual information based methods first proposed in 2012 and demonstrate that geometric features provide a robust information metric for camera-LiDAR extrinsic calibration. We demonstrate our proposed improvement using the KITTI and KITTI-360 fisheye data set.
\end{abstract}

\section{Introduction}
\label{sec:intro}

Modern autonomous vehicles and robots are powered by algorithms that fuse data from multiple sources in order to increase confidence and reduce decision making uncertainty. This algorithmic requirement for multiple independent observations of the environment requires that sensors be accurately co-registered in 3D space. Extrinsic calibration is the process of acquiring these co-registration parameters and is a fundamental building block of an autonomous system. 

Camera and LiDAR is a commonly used multi-sensor suite used in high level autonomy algorithms for self-driving cars and trucks. Accurately mapping image texture onto the LiDAR pointcloud or pointcloud texture (depth, speed or intensity) onto the camera image are two common functions that require the extrinsic calibration between camera and LiDAR. Even small errors in the rotational component of the extrinsic rigid body transform can introduce fatal errors into this data fusion process. For a review of image-pointcloud fusion applications, including extrinsic calibration see the work of Cui \etal \cite{Cui2020DeepLF}.

\begin{figure}
    \centering
    \includegraphics[width=0.48\linewidth]{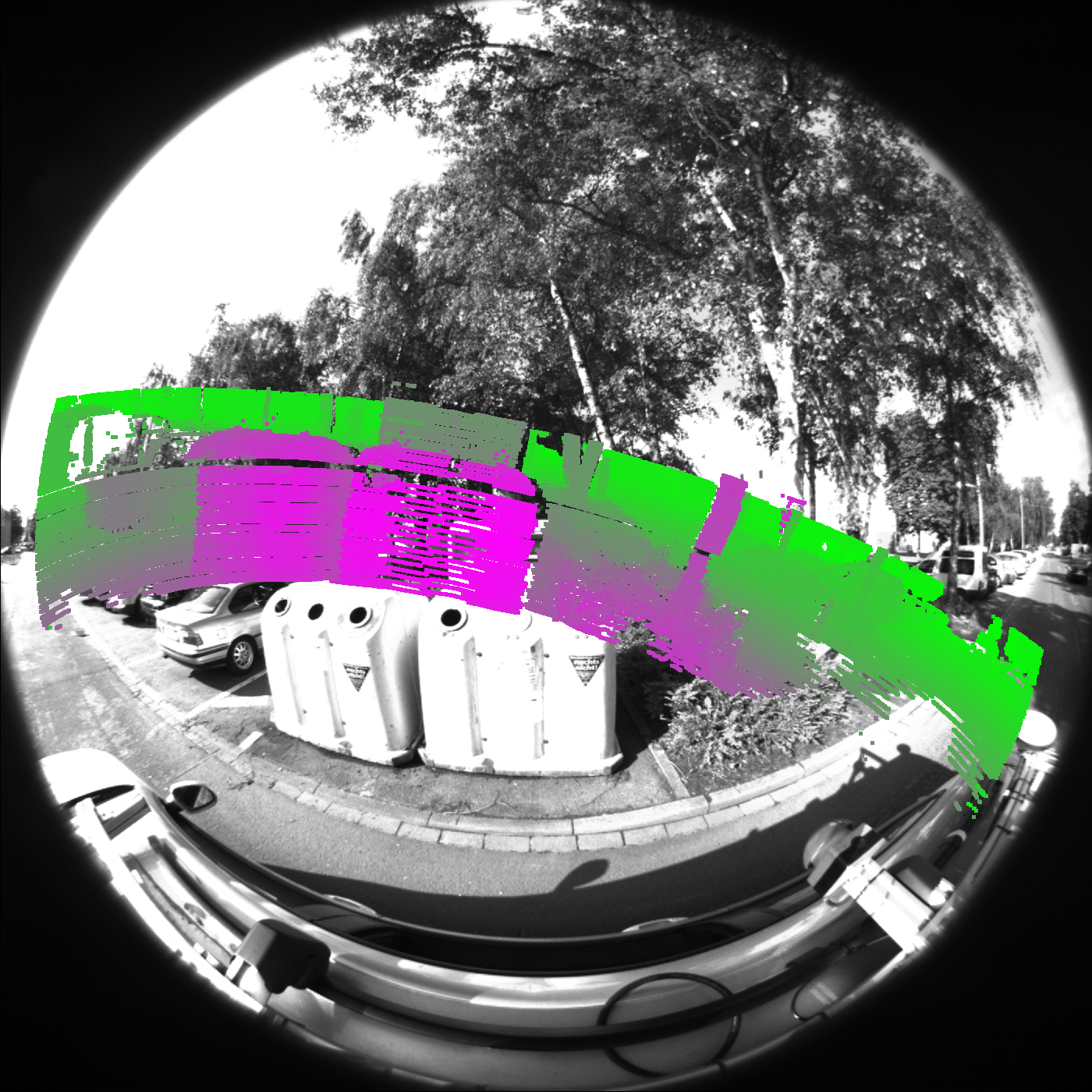}
    \includegraphics[width=0.48\linewidth]{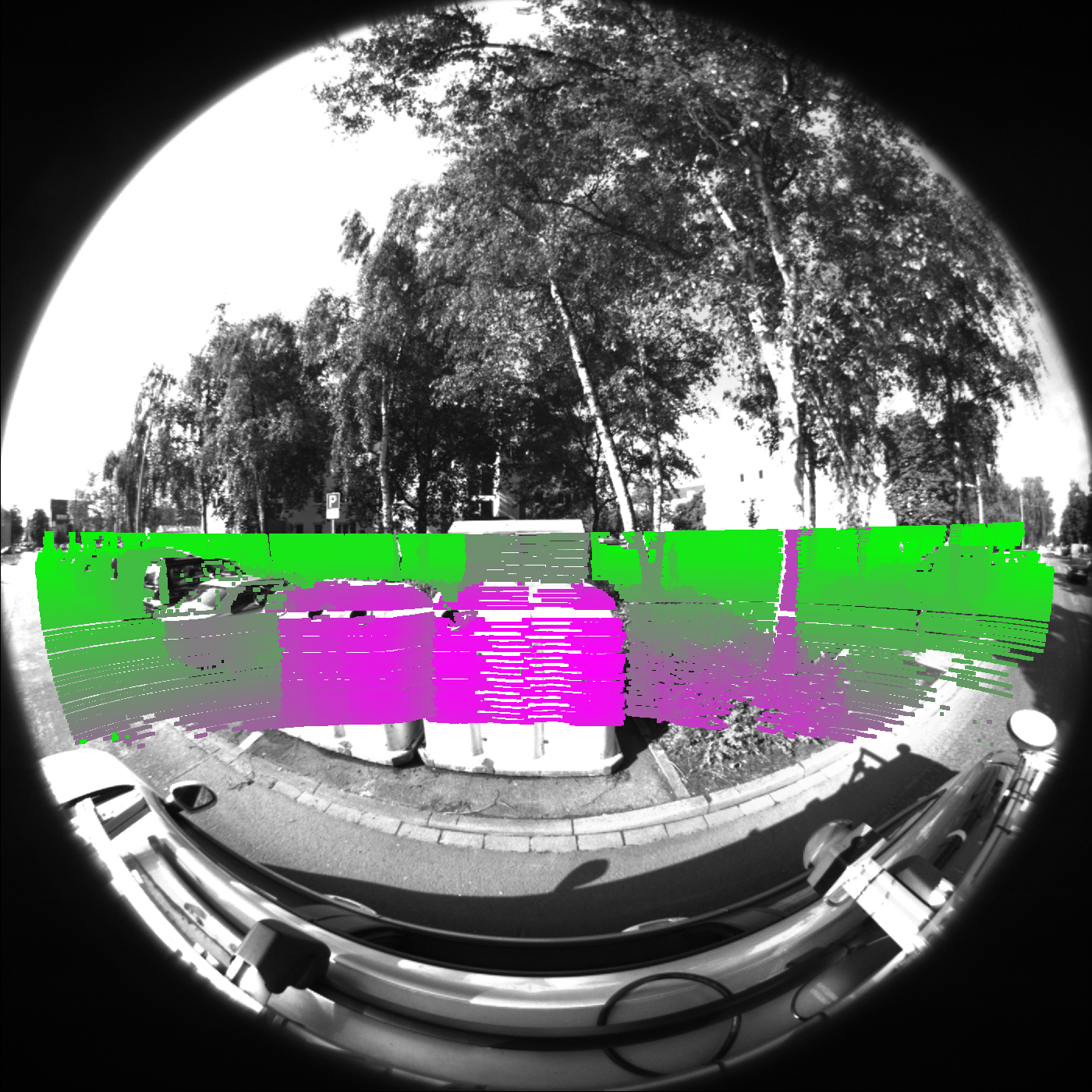}
    \caption{An accurate extrinsic calibration is required to fuse camera and LiDAR data together. A failed calibration results in a misaligned projection (left). Our target free calibration automatically provides the correct extrinsic calibration (right).}
    \label{fig:projection_results}
\end{figure}

Many extrinsic calibration algorithms, discussed in Section  \ref{sec:rel_work}, are challenging to automate. Therefore, it is common that extrinsic calibration is performed only once when a vehicle is factory commissioned. Naturally, this introduces the unrealistic constraint that the sensors will never move during the life of the vehicle. Online calibration methods, capable of calibrating sensors using only sensor data during the vehicles lifetime without requiring ground truth training data or artificial "calibration trajectories" are very relevant.

\begin{figure}
    \centering
    \setkeys{Gin}{width=\linewidth}
        \includegraphics{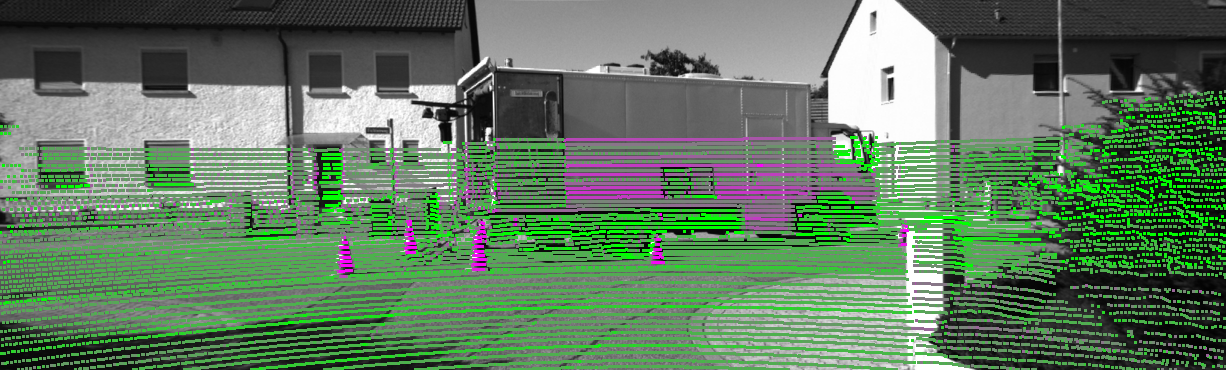}
        \includegraphics{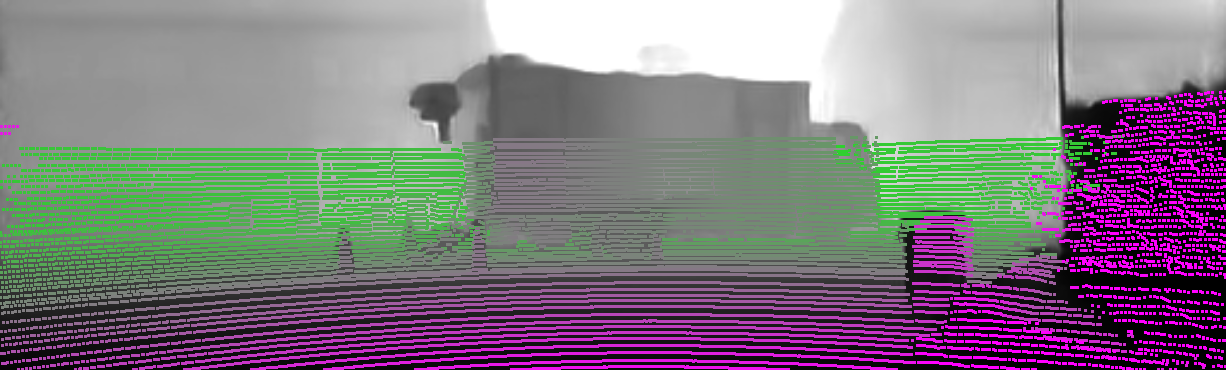}
    \caption{Comparison of intensity-to-intensity features (top) and depth-to-depth features (bottom). Depth features share significantly more information than intensity features. The depth estimate of the image (bottom) is provided by a self-supervised monocular depth estimation network that does not use any ground truth training data.}
    \label{fig:feature_comparison}
\end{figure}

Our contribution is an improvement to Mutual Information (MI) camera-LiDAR extrinsic calibration that dramatically increases robustness to initialization error, greatly increases precision and is accurate. We propose using a depth-to-depth (D2D) feature instead of the legacy intensity-to-intensity (I2I) feature (\cref{fig:feature_comparison}). To support our claim we provide a detailed justification and a comprehensive set of experiments on the KITTI and KITTI-360 data sets. 

\begin{figure*}
\center
    \includegraphics[width = \linewidth]{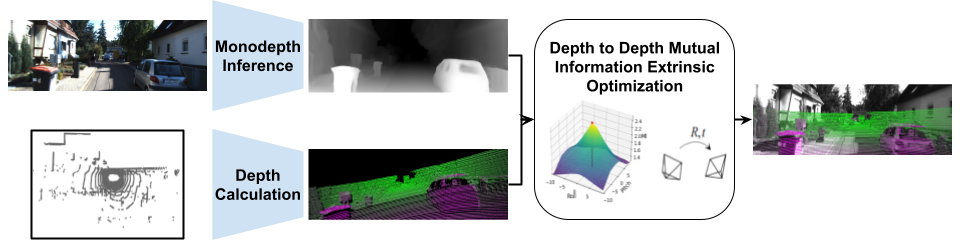} 
    \caption{Our proposed depth-to-depth mutual information extrinsic calibration workflow. A monocular depth estimation network generates depth predictions from the camera image and the LiDAR pointcloud is projected to an image with intensity proportional to depth. Finally, the extrinsic calibration is optimized by maximizing the mutual information between the two intramodal measurements.}
   \label{fig:coec_flow_chart}
\end{figure*}

\section{Related Work}
\label{sec:rel_work}

\subsection{Feature Extraction Based}
Feature extraction based extrinsic calibration algorithms can be broken down into three steps:

\begin{enumerate}
    \itemsep0em 
    \item Feature extraction
    \item Feature correspondence
    \item Reprojection error optimization
\end{enumerate}
   
Target-based and target-free feature based methods follow the same three steps. Target-based methods are provided strong correspondent features by a user configured fiducial target, a checkerboard for example. This is in contrast to target-free methods which are not provided strong correspondent features and must generate features which can be easily extracted and matched from inter-modal camera-LiDAR data.

\subsubsection{Target}
Zhang and Pless \cite{Zhang2004ExtrinsicCO} is the seminal work for target based offline camera-LiDAR calibration. In the image the checkerboard corners are detected and used to calculate the plane-normal and in the LiDAR pointcloud the points on the plane are segmented. This plane-normal and set of points on the plane is a geometric constraint which is used by a iterative nonlinear least-squares optimizer to solve for the camera-LiDAR extrinsic transformation. In follow-up work, the idea is expanded to new sensor models, initialization techniques and optimization constraints \cite{Unnikrishnan2005FastEC, Pandey2010ExtrinsicCO, Geiger2012AutomaticCA, Mirzaei20123DLI}.

\subsubsection{Target Free}
Scaramuzza \etal \cite{Scaramuzza2007ExtrinsicSC} removed the need for a calibration target by hand selecting features in the camera image and LiDAR pointcloud and solved a perspective-from-n-point problem followed by a non-linear refinement. Hand selected features are however tedious to collect and require an expert user.

A natural feature that exists in both camera and LiDAR data and can be semi-automatically extracted are edges. It is common that these inter-modal edges align and therefore metrics which measure edge magnitude and direction similarity can be used as a metric for camera-LiDAR extrinsic calibration. Multiple works use edge based metrics as an optimization constraint \cite{Bileschi2009FullyAC, Levinson2013AutomaticOC, Taylor2013AutomaticCO, Taylor2015MultiModalSC, MuozBan2020TargetlessCC}. Edge based methods are very promising for autonomous vehicle calibration because they require neither an operator or fiducial target but are highly sensitive to initialization error. 

Learning based methods starting with the work of Schneider \etal \cite{Schneider2017RegNetMS} cast all three steps: feature extraction, matching and optimization into a neural network learning problem. These methods require minimal feature engineering and no specific calibration targets. Work to improve the loss function and network architecture has made progress but has not yet solved the fundamental domain gap between camera and LiDAR data \cite{Iyer2018CalibNetGS, Lv2021LCCNetLA, Zhao2021CalibDNNMS, Zhang2022EnhancedLL, Sun2023ATOPAA, Jeon2022EFGHNetAV}. Notably, these methods require ground truth training data, in the form of a ground truth extrinsic calibration or semantic labels. This data is impossible to collect at scale for real world applications. Our proposed D2D optimization algorithm does not require any ground truth labeled data and is therefore easily used, in contrast to the above listed black-box over fit neural networks. 

\subsection{Odometry Based}
Robot hand-eye calibration using structure from motion was used for the extrinsic calibration of a camera and a mobile platform with known ego-motion in the work of Andreff \etal \cite{Andreff2001RobotHC}. With the advent of LiDAR based ego-motion algorithms, the natural extension of this was to calculate the extrinsic calibration between a camera and a LiDAR with calculated ego-motion \cite{Taylor2015MotionbasedCO}. The value of the extrinsic calibration is regressed from the camera and LiDAR trajectory just as in the hand-eye calibration process. Multiple works present examples of hand-eye camera-LiDAR extrinsic calibration algorithms \cite{Chien2016VisualOD, Ishikawa2018LiDARAC, ZuigaNol2019AutomaticME}. These methods are promising for autonomous vehicle calibration because they do not require an operator, fiducial target or even overlapping sensor field of view. A flaw of odometry based methods is that most vehicle motion rotates around only a single axis (yaw) and therefore the roll and pitch are often now observable without specially designed calibration trajectories.

\subsection{Mutual Information Based}
\begin{figure}[!htb]
\center
  \subcaptionbox{De-calibrated (left, MI=0.39) and calibrated (right, MI=0.47) intensity-to-intensity histograms.}{
    \includegraphics[width = 0.45\linewidth]{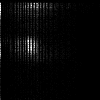} 
    \includegraphics[width =  0.45\linewidth]{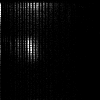}}\quad
  
  \subcaptionbox{De-calibrated (left, MI=2.40) and calibrated (right, MI=3.05) depth-to-depth histograms.}{
    \includegraphics[width =  0.45\linewidth]{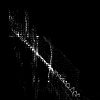}
    \includegraphics[width =  0.45\linewidth]{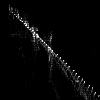}}\quad

  \caption{Maximizing mutual information is equivalent to sharpening the joint feature histogram. The depth feature histogram (bottom row) is more structured and sharpens significantly more than the intensity feature histogram (top row) for the same re-calibration. Our proposed depth features are directly correlated therefore we see a linear relationship between the camera and LiDAR feature.}
   \label{fig:histograms}
\end{figure}

Mutual information, also called relative entropy, is an established generic measure of the statistical dependence between two random variables \cite{Vajda1989TheoryOS, Cover1991ElementsOI}. Maximizing the amount of mutual information, which can also  be interpreted as "sharpening" the joint histogram of the distributions (Fig. \ref{fig:histograms}), is the criterion for MI based extrinsic calibration algorithms. This approach is based on the assumption that at the correct extrinsic calibration the MI of the distributions is maximized. Mutual information, which makes no assumption about the distribution's shape, form or dependency has been found to be a highly effective metric for extrinsic calibration of inter-modal medical images \cite{Viola1995AlignmentBM, Wells1996MultimodalVR, Maes1997MultimodalityIR}.

\subsubsection{Robotics}
Mutual information extrinsic calibration of a camera-LiDAR pair was first applied to aerial survey data \cite{Mastin2009AutomaticRO}. Instead of maximizing the MI however, the entropy of the image and LiDAR features is assumed constant for small perturbations, and the joint entropy $H(X,Y)$ is minimized. The camera image's intensity and the LiDAR's elevation and probability of detection was used to construct two joint feature entropies that were minimized. Ground based surveys also require accurate extrinsic camera-LiDAR calibration. Taylor and Nieto \cite{Nieto2012AMI} calibrated a stationary hyper-spectral camera and 3D LiDAR using the MI between image intensity and LiDAR pointcloud point normal features.

Pandey \etal \cite{Pandey2012AutomaticTE} applied the concept to an automotive camera-LiDAR data set, calculating the MI between grayscale image intensity and LiDAR intensity. Interestingly, using pointcloud point depth as the feature, instead of intensity, was proposed but not evaluated. This was followed up by a 2015 paper \cite{Pandey2015AutomaticEC} which expanded the idea to multiple new sensing modalities and environments. In both papers it was noted that environments with features close to the sensors and with high correlated image intensity and LiDAR intensity (i.e. metal signs, road paint or cars) result in better calibrations. Our contribution made in this paper is a direct successor of and largely inspired by this work.

\subsection{Our Contribution}
We propose maximizing geometric feature mutual information as the optimization metric for camera-LiDAR extrinsic calibration. Advances in monocular depth estimation \cite{Godard2018DiggingIS, Ranftl2019TowardsRM} have enabled the extraction of geometric information from camera images without the need for ground truth labeled training data. Our method takes advantage of this fact and calculates the MI between the inferred camera depth image and the LiDAR pointcloud's inherent geometric depth feature. 

The result is a target free camera-LiDAR calibration algorithm that requires no ground-truth training data or hand-crafted features and is explainable. Our contribution extends naturally to other dense depth sensors, such as structured light sensors, time-of-flight sensors and stereo camera rigs, where image-LiDAR correspondences can easily be calculated with a camera model.

\section{Theory}

\begin{figure}
    \centering
    \setkeys{Gin}{width=1\linewidth}
    \subcaptionbox{I2I}{\includegraphics{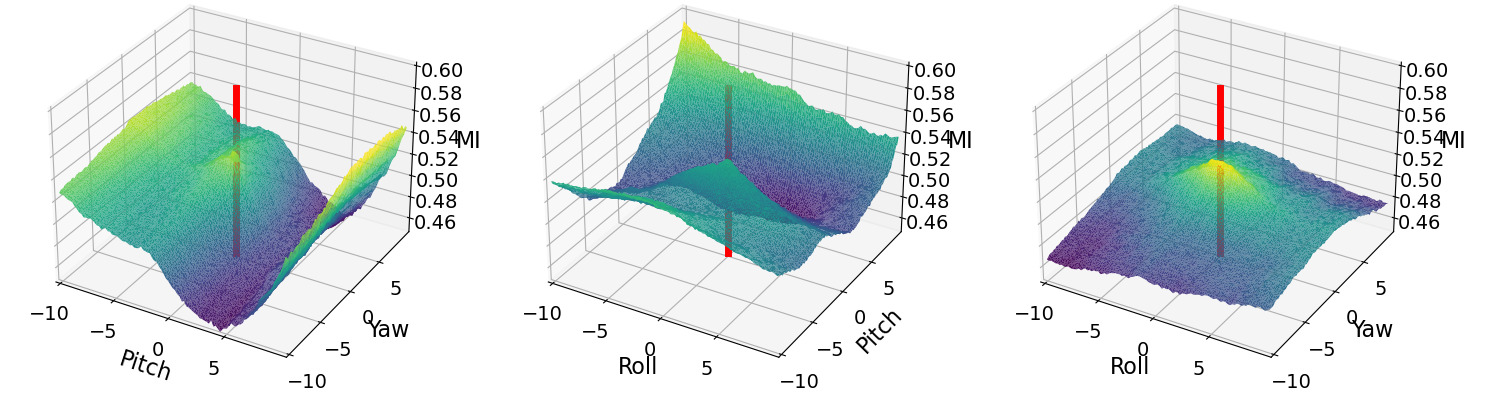}}
    \subcaptionbox{D2D}{\includegraphics{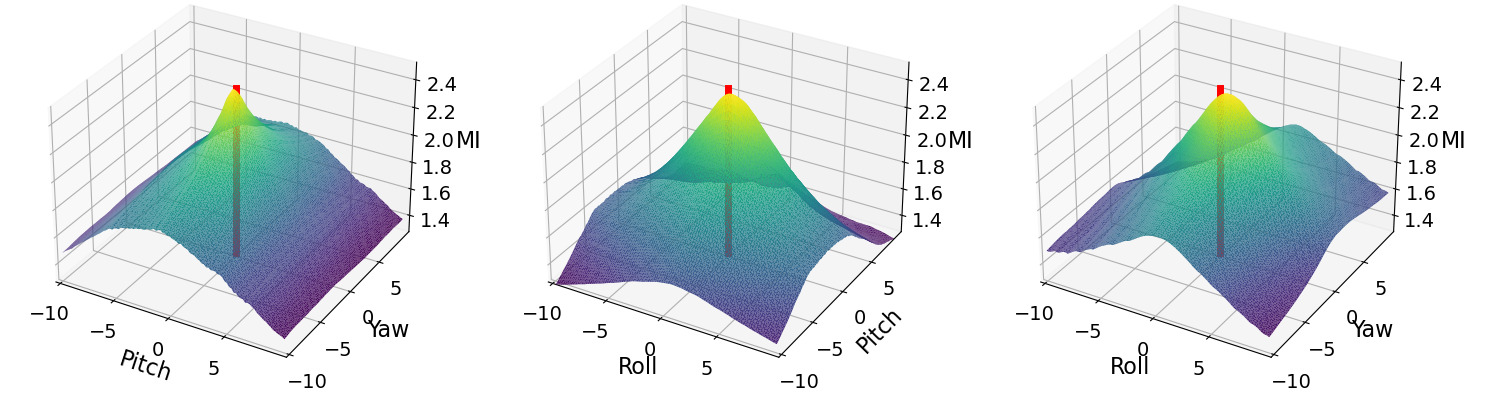}}
    \caption{Mutual information surface plots of (i) classic I2I, and (ii) our proposed D2D feature from 25 frames. The D2D MI is clearly more suitable for optimization over a larger range of errors than I2I.}
    \label{fig:mi-surface-plots}
\end{figure}

\subsection{Statistical Theory}
\label{subsec:stattheory}

Statistical dependence refers to the relationship or association between two or more variables. If two variables are statistically independent, the distribution of one remains unchanged regardless of the knowledge of the other. Conversely, if two variables are dependent, knowledge of the distribution of one provides total information about the distribution of the other. These two extremes represent the limits of association, where independent variables share no information and dependent variables share complete information.

Using the definition of entropy $H = -\sum_{i} p_{i} \log{p_{i}}$ the mutual information of two variables can be written in terms of the entropy of the marginal distributions and the joint distributions with

\begin{equation} 
    \label{eqn:mi}
    I(X,Y) = H(X) + H(Y) - H(X,Y).
\end{equation}

\subsection{Measurement Theory}
Camera and LiDAR both sense the surrounding environment. A camera passively captures the visible light texture and the LiDAR actively senses non-visible light texture which is then processed into a 3D position and intensity. Modeling how camera and LiDAR sensors measure intensity is notoriously challenging. For image sensors the amount of light measured is highly dependent on external lighting conditions, object color and reflectance properties. Three variables which in commonly encountered environments are difficult to efficiently and accurately constrain. Measured LiDAR intensity is similarly difficult to model and is highly dependent on surface finish, color, and angle of incidence. Three variables which, like modeling camera intensity, are also nearly impossible to efficiently and accurately constrain. 

Therefore the assumption that camera and LiDAR intensities are correlated is problematic. Practical examples of this problem are commonly found in autonomous driving data. Where for example two cars, one dark red and one black can have similar camera intensity values but completely different LiDAR intensity values. Or two road signs, one dark blue and one white that have nearly the same LiDAR intensity values but vastly different camera intensity values.

With the advent of monocular depth estimation algorithms we can now consider a third virtual depth sensor that directly maps image intensity to depth. Such algorithms are robust and can produce estimates of scene geometry which are accurate enough for our geometric MI calibration metric \cite{Godard2018DiggingIS, Ranftl2019TowardsRM}. LiDAR sensors have also experienced large improvements in their ability to measure scene geometry and are able to reliably estimate 3D position to within several centimeters at ranges over 100m.  

The ability of both a virtual depth sensor and LiDAR to measure 3D geometry is the foundation of the extrinsic calibration algorithm presented in this paper.

\section{Method}
The foundation of our method is the mutual information based extrinsic calibration of a camera and LiDAR. We improve on the previous MI-based calibration approaches of Mastin \etal \cite{Mastin2009AutomaticRO} and Pandey \etal \cite{Pandey2015AutomaticEC} by using highly correlated depth features instead of poorly correlated intensity features.

\subsection{Mathematical Formulation}
\label{sec:math_form}
In the following, we posit the algorithm in terms of the LiDAR coordinate frame $\boldsymbol{O}_{L}$ and the camera coordinate frame $\boldsymbol{O}_{C}$. We denote the rigid transformation that transforms a point $\boldsymbol{p}_L$ in coordinate frame $\boldsymbol{O}_{L}$ to a corresponding point $\boldsymbol{p}_C$ in coordinate frame $\boldsymbol{O}_{C}$ by 

\begin{equation}
  \sideset{^C}{_L}{\mathop{\mathbf{T}}} = (\sideset{^C}{_L}{\mathop{\mathbf{R}}}, \sideset{^C}{_L}{\mathop{\mathbf{t}}}) \in \mathbb{R}^{4\times4}
\end{equation}

where $\sideset{^C}{_L}{\mathop{\mathbf{R}}} \in ~\mathrm{SO}(3)$ and $\sideset{^C}{_L}{\mathop{\mathbf{t}}} \in \mathbb{R}^{3}$ are the rotational and translational part of the transformation. Using homogeneous coordinates, a 3D point $\bm{p}_L$ in the LiDAR coordinate frame $\bm{O}_{L}$ can be transformed to a point $\bm{p}_C$ in the camera coordinate frame $\bm{O}_{C}$ with $\bm{p}_C = \sideset{^C}{_L}{\mathop{\mathbf{T}}} \cdot \bm{p}_L$.

Using the pinhole camera model for simplicity, a homogeneous point $\boldsymbol{p}_C$ can be further transformed to a point $\boldsymbol{p}_{CO}$ in the 2D camera optical coordinate frame $\boldsymbol{O}_{CO}$ with

\begin{equation}  
    \boldsymbol{p}_{CO} = \boldsymbol{P}\cdot \boldsymbol{p}_C
\end{equation}

where $\boldsymbol{P} = \boldsymbol{K}[\boldsymbol{I}|\boldsymbol{0}] \in \mathbb{R}^{3\times4}$ is the projective camera matrix parameterized by the calibration matrix $\boldsymbol{K}$. It should be emphasized that the algorithm is generic to the selected camera model, and in this paper we use both the pinhole model for the KITTI experiment and double sphere model for the KITTI-360 experiment.

\begin{figure}[h]
    \centering
    \setkeys{Gin}{width=\linewidth}
    \includegraphics{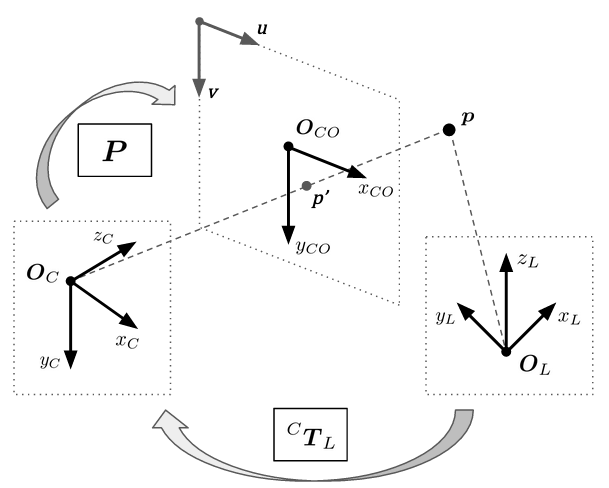}
    \caption{Using the rigid transformation $\sideset{^C}{_L}{\mathop{\mathbf{T}}}$ and the projective transformation $\bm{P}$, a 3D point $\bm{p}$ in the LiDAR frame can be transformed to a 2D pixel in the camera image.}
    \label{fig:tf}
\end{figure}

Using both transformations, $\sideset{^C}{_L}{\mathop{\mathbf{T}}}$ and $\bm{P}$, a 3D point $\boldsymbol{p}_L$ measured in the LiDAR coordinate frame can be projected to a 2D point $\boldsymbol{p}'$ in the camera optical coordinate frame. The projected point can then be represented by its pixel coordinates $(\boldsymbol{u}', \boldsymbol{v}')$ in the image frame. \cref{fig:tf} illustrates these transformations.

\begin{figure}
    \centering
    \subcaptionbox{I2I}{
        \includegraphics[width =  0.48\linewidth]{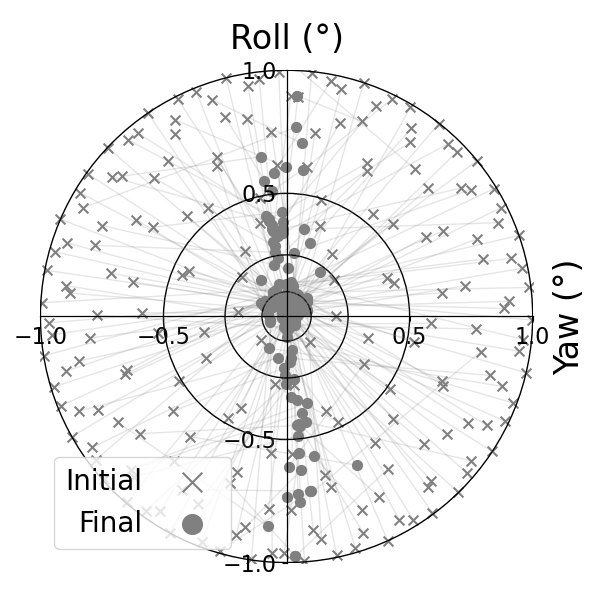}}
    \subcaptionbox{D2D}{
        \includegraphics[width =  0.48\linewidth]{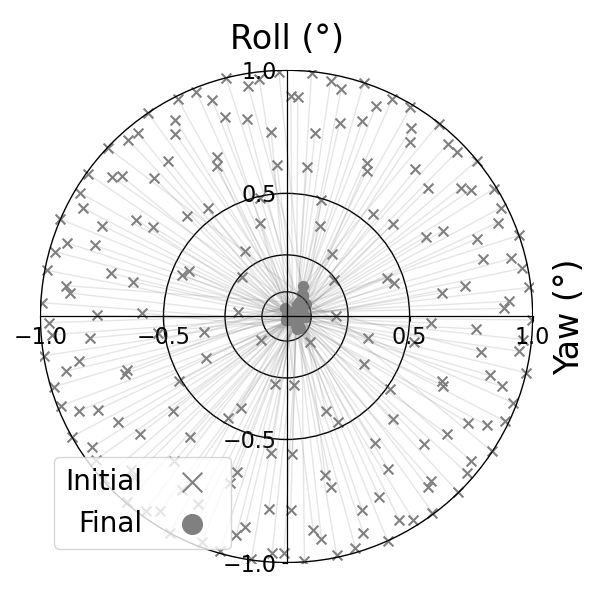}}\quad
    \caption{We introduce the Bull's Eye plot. It is designed such that the ground truth extrinsic parameter value is at the center of the plot. A single execution of the algorithm is represented by two connected points, the initial perturbed extrinsic value marked with an "x" and the final optimized value marked with a dot. Plotted here are the 200 executions of the $1^{\circ}$ 3 DoF experiment. The improvement in accuracy and precision of D2D compared to I2I is clear.}
    \label{fig:bullseye}
\end{figure}

\subsection{Algorithm}
We define $\mathcal{F} = \{(I_1, P_1), (I_2, P_2), ..., (I_{N}, P_{N})\}$ as a set of $N$ time synchronized image pointcloud pairs. Each image $I_i$ is a grid of pixels $i_{0},i_{1},...,i_{j}$ measuring intensity. Each pointcloud $P_i$ is a set of $K$ 3D points $\boldsymbol{p}_{1}, \boldsymbol{p}_{2}, ...,\boldsymbol{p}_{K}$ which can include additional measurement attributes such as intensity. In the first step a depth map $D_i$ is inferred for each image by a monocular depth estimation network to generate a set of depth maps $\mathcal{D} = \{D_1, D_2, ..., D_{N}\}$. An initial guess of the extrinsic parameters $\boldsymbol{\Theta}_0 = [\theta_{x}, \theta_{y}, \theta_{z}, t_{x}, t_{y}, t_{z}]^T$ is used to generate the transformation 

\begin{align}
    \sideset{^C}{_L^0}{\mathop{\mathbf{T}}} = [\boldsymbol{R}|\boldsymbol{t}]
\end{align}

where $\boldsymbol{R} = \boldsymbol{R}_x(\theta_x) \cdot \boldsymbol{R}_y(\theta_y) \cdot \boldsymbol{R}_z(\theta_z) \in \mathbb{R}^{3 \times 3}$ represents a chained rotation around the three canonical axes and $\boldsymbol{t} \in \mathbb{R}^{3}$ the translation vector. Using this transformation together with the camera projection, each pointcloud $P_i$ is projected into the image frame as explained in \cref{sec:math_form}. Points that are outside the camera's field of view are omitted. 

The resulting set $U_{i}^P = \{ (u,v)_{1}, (u,v)_{2},...,(u,v)_{M}\}$ with $M\leq K$ contains the pixel coordinates of all points from $P_i$ that have been successfully projected into the image. From the corresponding 3D points we directly get a set of depth features $f^P = \{d_1^P, d_2^P, ..., d_M^P\}$ for pointcloud $P_i$ using the euclidean distance $d_i^P=\|\boldsymbol{p}_{L_i}\|_2$ of each point to the LiDAR frame origin. Likewise, we extract a set of matching depth features $f^D = \{d_1^D, d_2^D, ..., d_M^D\}$ from depth map $D_i$ using the depth values at the pixel coordinates $U_{i}^P$. This step is the point in the algorithm where correspondence is directly established between camera and LiDAR features.

To calculate the mutual information between the feature sets, we approximate their marginal and joint distributions $p(f^P)$, $p(f^D)$ and $p(f^P, f^D)$ using normalized histograms (\cref{fig:histograms}). We express the MI value as a function of the image pointcloud pair $(P_i, D_i)$ and the extrinsic parameters $\boldsymbol{\Theta}$ with

\begin{align}
  MI(P_i, D_i, \boldsymbol{\Theta}) = H(f_{{\boldsymbol{\Theta}}}^P) + H(f_{{\boldsymbol{\Theta}}}^D)  - H(f_{{\boldsymbol{\Theta}}}^P,f_{{\boldsymbol{\Theta}}}^D).
\end{align}
The above expression is based on \cref{eqn:mi} introduced in \cref{subsec:stattheory}.

To get an accurate estimate of the MI value including all image pointcloud pairs we average over the whole set $\mathcal{F}$ yielding the final value
\begin{align}
    \label{eq:mi_avg}
  MI(P_i, D_i, \boldsymbol{\Theta}) = \frac{1}{N}\sum_{i=1}^N MI(P_i, D_i, \boldsymbol{\Theta}).
\end{align}
Using \cref{eq:mi_avg} as an objective function, we can apply a suitable iterative algorithm to solve for a set of parameters $\boldsymbol{\hat \Theta}$ that maximize the average MI value given a set of image pointcloud pairs. Due to the perspective projection of the 3D points it is not possible to analytically calculate the derivative of the objective function with respect to the parameters $\boldsymbol{\Theta}$. In contrast to Pandey \etal \cite{Pandey2015AutomaticEC} we therefore apply Powell's BOBYQA algorithm \cite{bobyqa} which efficiently solves the bounded optimization problem without derivatives using quadratic approximation. Algorithm \ref{alg:coec} shows pseudocode for the procedure.

\begin{algorithm}
    \caption{D2D MI Extrinsic Calibration}
    \label{alg:coec}
    \begin{algorithmic}
       \State \textbf{Input:} Initial guess $\boldsymbol{\Theta}_0$, set of $N$ image  pointcloud pairs  \newline\hspace*{2.8em} $\mathcal{F} = \{(I_1, P_1), (I_2, P_2), ..., (I_{N}, P_{N})\}$
        \State \textbf{Ouput:} Optimized parameter $\boldsymbol{\hat \Theta}$
        \State $D_{1...N} \gets$  \Call{ExtractDepthFeatures}{\textit{$I_{1...N}$}}  
        \State $\boldsymbol{\hat \Theta} \gets$  \Call{BOBYQA}{\Call{CalcMI}{$\boldsymbol{\Theta}_0$, $D_{1...N}$, $P_{1...N}$}}  
        \State \Return $\boldsymbol{\hat \Theta}$

        \Procedure{CalcMI}{$\boldsymbol{\Theta}$, $D_{1...N}$, $P_{1...N}$}
        \State $MI =0$
        \State $\sideset{^C}{_L}{\mathop{\mathbf{T}}} = [\boldsymbol{R}|\boldsymbol{t}] \gets \boldsymbol{\Theta}$
        \For{$i \gets 1$ to $N$}  
        \State $f^P, f^D \gets \Call{GetMatches}{\sideset{^C}{_L}{\mathop{\mathbf{T}}}, D_{1...N}, P_{1...N}}$
            \State $MI \mathrel{+}= H(f^P) + H(f^D)  - H(f^P,f^D)$
        \EndFor
        \State \Return $\frac{MI}{N}$
    \EndProcedure
    \end{algorithmic}
\end{algorithm}

\section{Experiments}
Two experiments were performed using two different camera models in two data sets. The experiments are a 3 Degrees of Freedom (DoF) rotation optimization only experiment and a 6 DoF rotation and translation optimization experiment. The two data sequences used are from KITTI where we use a pinhole camera model and KITTI-360 where we use a double sphere camera model and use the raw unrectified $180^{\circ}$ FoV images directly. For both the KITTI and KITTI-360 sequences only 25 frames out of the sequence, sampled uniformly from the entire length of the sequence were used.

In both experiments, the ground truth value of the extrinsic calibration is perturbed by applying a transformation to the LiDAR pointcloud. For each error range ($1^{\circ}$, $2^{\circ}$, etc.) 200 points are generated on a unit sphere using the Fibonacci sphere algorithm. The components of each of the 200 unit vectors are multiplied by the selected error range. The same is done for the translation perturbation in the 6 DoF experiment.

In the interest of reproducibility, and because monocular depth estimation itself is not a focus of this work, we use two publicly available networks. For the KITTI sequence we use the self-supervised trained FeatDepth network \cite{Shu2020Featdepth} and for KITTI-360 we use the partially-supervised MiDaS network \cite{Ranftl2022MiDaS}.

A "hit" metric was defined to remove executions that are obviously catastrophic failures from the statistic calculations. A optimization is considered to have "hit" (been successful) if the magnitude of the optimized rotation parameters is less than $0.5^{\circ}$ and of the translation parameters less than $20$cm.

\begin{table}
\begin{center}
\begin{tabular}{ |l|c|c| } 
    \hline
        Data & KITTI & KITTI-360 \\ 
    \hline \hline
        Date & 09/30/2011 & 05/28/2013 \\ 
        Drive & 0018 & 0000 \\ 
        Frames & 2.762 & 11,518 \\
        Camera & image\_02 & image\_03 \\
        Depth & FeatDepth & MiDaS \\
    \hline
\end{tabular}
\end{center}
\caption{Description of the two sequences used in the experiments.}
\label{table:data_set}
\end{table}

\begin{figure*}
    \includegraphics[width=.24\textwidth]{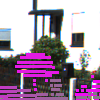}\hfill
    \includegraphics[width=.24\textwidth]{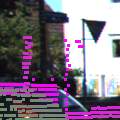}\hfill
    \includegraphics[width=.24\textwidth]{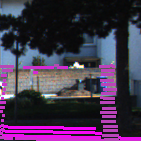}\hfill
    \includegraphics[width=.24\textwidth]{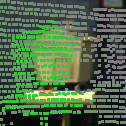}
    \\[\smallskipamount]
    \includegraphics[width=.24\textwidth]{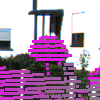}\hfill
    \includegraphics[width=.24\textwidth]{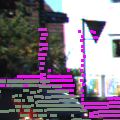}\hfill
    \includegraphics[width=.24\textwidth]{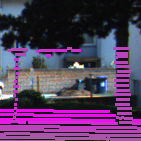}\hfill
    \includegraphics[width=.24\textwidth]{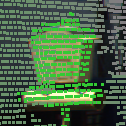}
    \\[\smallskipamount]
    \includegraphics[width=.24\textwidth]{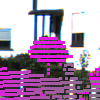}\hfill
    \includegraphics[width=.24\textwidth]{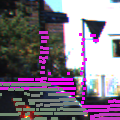}\hfill
    \includegraphics[width=.24\textwidth]{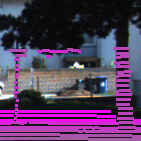}\hfill
    \includegraphics[width=.24\textwidth]{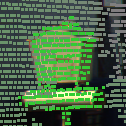}
    \caption{Qualitative results: projection before optimization (top), projection after proposed D2D optimization (middle), and projection with ground truth calibration (bottom).}\label{fig:foobar}
\end{figure*}

\subsection{Rotation Only Experiment}
In this experiment we model the error as a rotation of the LiDAR, however camera rotation can also be considered. This experiment reflects a real world use case where there is strong prior on the translation component of the extrinsic calibration. Such as in autonomous vehicles or robotic systems in which the translation is known from design drawings or a pre-exisiting factory calibration.

The results of this experiment are shown in Table \ref{table:results_rotation}. The superior hit rate of D2D is clear. At a $10^{\circ}$ error, I2I hits none of the KITTI executions and only 0.5\% of the KITTI-360 executions, D2D however hits 96.5\% and 100\% of executions respectively. Even up to an error of $20^{\circ}$ D2D hits 50.5\% and 89\% of KITTI and KITTI-360 executions. The increased hit rate of KITTI-360 is due to the fact the LiDAR pointcloud never goes of out the FoV because of the wide FoV fisheye lens.

The precision of D2D is superior to I2I. At all four error levels, for both KITTI and KITTI-360, the mean value of each respective rotation parameter varies by less than $0.02^{\circ}$. There is no comparable consistency for the I2I optimization. 

The accuracy of I2I is marginally better, when it hits, for the KITTI sequence but the same for the KITTI-360 sequences. For example, for the $1^{\circ}$ KITTI experiment the magnitude of the I2I and D2D error is $0.16^{\circ}$ and $0.29^{\circ}$ respectively. For the KITTI-360 sequence the magnitude of the I2I and D2D error is $0.38^{\circ}$ and $0.37^{\circ}$ respectively.

We propose using Bull's Eye plots, shown in \cref{fig:bullseye} to easily characterize and communicate the accuracy of 3 DoF extrinsic calibration optimization algorithms.

\begin{table*}
\begin{center}
    \resizebox{\textwidth}{!}{\begin{tabular}{|l||c|c|c|c||c|c|c|c||c|c|c|c||c|c|c|c|}
    \hline
     Error & \multicolumn{4}{c||}{ $1^{\circ}$} & \multicolumn{4}{c||}{ $2^{\circ}$} & \multicolumn{4}{c||}{ $10^{\circ}$} & \multicolumn{4}{c|}{ $20^{\circ}$} \\
     \hline
     Data & \multicolumn{2}{c|}{KITTI} & \multicolumn{2}{c||}{KITTI-360} & \multicolumn{2}{c|}{KITTI} & \multicolumn{2}{c||}{KITTI-360} & \multicolumn{2}{c|}{KITTI} & \multicolumn{2}{c||}{KITTI-360} & \multicolumn{2}{c|}{KITTI} & \multicolumn{2}{c|}{KITTI-360} \\
     \hline
     Method & I2I & D2D & I2I & D2D & I2I & D2D & I2I & D2D & I2I & D2D & I2I & D2D & I2I & D2D & I2I & D2D \\
     \hline \hline
     Hit (\%) & 61 & \textbf{100} & 46.5 & \textbf{100} & 12.5 & \textbf{99.5} & 27.5 & \textbf{100} & 0 & \textbf{96.5} & 0.5 & \textbf{100} & 0 & \textbf{50.5} & 0 & \textbf{89} \\
     \hline
     \hline
      
     \shortstack{$\theta_{R}$ ($^{\circ}$) \\ \vspace{1pt}} & 
     \shortstack{-0.07 \\ $\pm$ 0.19} & \shortstack{0.26 \\ $\pm$ 0.05} & \shortstack{-0.35 \\ $\pm$ 0.04} & \shortstack{0.30 \\ $\pm$ 0.02} & 
        \shortstack{-0.10 \\ $\pm$ 0.11} & \shortstack{0.26 \\ $\pm$ 0.05}  &  \shortstack{-0.35 \\ $\pm$ 0.05} & \shortstack{0.30 \\ $\pm$ 0.02} &  
            \shortstack{- \\ -} & \shortstack{0.27 \\ $\pm$ 0.05}  &  \shortstack{-0.34 \\ $\pm$ -} & \shortstack{0.30 \\ $\pm$ 0.02} & 
                \shortstack{- \\ -} & \shortstack{0.27 \\ $\pm$ 0.05}  & \shortstack{- \\ -} & \shortstack{0.30 \\ $\pm$ 0.02} \\
     \hline
     
     \shortstack{$\theta_{P}$ ($^{\circ}$) \\ \vspace{1pt}} & 
     \shortstack{-0.13 \\ $\pm$ 0.08} &  \shortstack{-0.12 \\ $\pm$ 0.07} & \shortstack{0.11 \\ $\pm$ 0.10} & \shortstack{-0.03 \\ $\pm$ 0.02} & 
        \shortstack{-0.10 \\ $\pm$ 0.04} & \shortstack{ -0.11 \\ $\pm$ 0.07}  & \shortstack{0.07 \\ $\pm$ 0.10} & \shortstack{-0.03 \\ $\pm$ 0.02} & 
            \shortstack{- \\ -} & \shortstack{-0.11 \\ $\pm$0.08 }  & \shortstack{0.08 \\ $\pm$ -} & \shortstack{-0.03 \\ $\pm$ 0.02} & 
                \shortstack{- \\ -} & \shortstack{-0.13 \\ $\pm$ 0.06}  & \shortstack{- \\ -} & \shortstack{-0.03 \\ $\pm$ 0.02} \\
     \hline
     
     \shortstack{$\theta_{Y}$ ($^{\circ}$) \\ \vspace{1pt}} & 
     \shortstack{0.07 \\ $\pm$ 0.07} & \shortstack{0.04 \\ $\pm$ 0.02}  &   \shortstack{-0.08 \\ $\pm$ 0.08} & \shortstack{0.21 \\ $\pm$ 0.04} & 
        \shortstack{0.06 \\ $\pm$ 0.04} & \shortstack{0.04 \\ $\pm$ 0.02}  & \shortstack{-0.09 \\ $\pm$ 0.08} & \shortstack{0.21 \\ $\pm$ 0.03} & 
            \shortstack{- \\ -} & \shortstack{0.04 \\ $\pm$ 0.02}  & \shortstack{-0.11 \\ $\pm$ -} & \shortstack{0.21 \\ $\pm$ 0.04} & 
                \shortstack{- \\ -} & \shortstack{0.03 \\ $\pm$ 0.02}  & \shortstack{- \\ -} & \shortstack{0.21 \\ $\pm$ 0.03} \\
     \hline
    \end{tabular}}
\end{center}
\caption{Rotation only extrinsic calibration optimization results using I2I and D2D mutual information at four different error levels. Our proposed D2D method can converge to the true value of the extrinsic calibration from over a $20^{\circ}$ error. D2D optimization has significantly more "hits" and is more precise than I2I optimization. Statistics are calculated only for optimizations that hit.}
\label{table:results_rotation}
\end{table*}

\begin{table*}
\tiny
\begin{center}
    \resizebox{\textwidth}{!}{\begin{tabular}{|l||c|c|c|c||c|c|c|c||c|c|c|c||c|c|c|c|}
    \hline
     Error & \multicolumn{4}{c||}{ $0.5^{\circ}$, $25$cm} & \multicolumn{4}{c||}{ $1^{\circ}$, $25$cm} & \multicolumn{4}{c||}{ $0.5^{\circ}$, $50$cm} & \multicolumn{4}{c|}{ $1^{\circ}$, $50$cm} \\
     \hline
     Data & \multicolumn{2}{c|}{KITTI} & \multicolumn{2}{c||}{KITTI-360} & \multicolumn{2}{c|}{KITTI} & \multicolumn{2}{c||}{KITTI-360} & \multicolumn{2}{c|}{KITTI} & \multicolumn{2}{c||}{KITTI-360} & \multicolumn{2}{c|}{KITTI} & \multicolumn{2}{c|}{KITTI-360} \\
     \hline
     Method & I2I & D2D & I2I & D2D & I2I & D2D & I2I & D2D & I2I & D2D & I2I & D2D & I2I & D2D & I2I & D2D \\
     \hline \hline
     Hit (\%) &
        4 & \textbf{84.5} & 26 & \textbf{71.5} &
            0.5 & \textbf{51.5} & 9 & \textbf{49} & 
                9.5 & \textbf{88} & 22 & \textbf{76.5} &
                    2 & \textbf{40.5} & 3 & \textbf{40.5} \\
     \hline
     \hline
     
    \shortstack{$\theta_{R}$ ($^{\circ}$) \\ \vspace{1pt}} &
        \shortstack{-0.05 \\ $\pm$ 0.14} & \shortstack{0.15 \\ $\pm$ 0.09} & \shortstack{-0.11 \\ $\pm$ 0.18} & \shortstack{-0.06 \\ $\pm$ 0.25} &
            \shortstack{-0.06 \\ $\pm$ -} & \shortstack{0.18 \\ $\pm$ 0.11} & \shortstack{-0.04 \\ $\pm$ 0.20} & \shortstack{-0.02 \\ $\pm$ 0.23} & 
                \shortstack{-0.08 \\ $\pm$ 0.16} & \shortstack{0.15 \\ $\pm$ 0.10} & \shortstack{-0.10 \\ $\pm$ 0.21} & \shortstack{0.07\\ $\pm$ 0.23} &
                    \shortstack{-0.04 \\ $\pm$ 0.09} & \shortstack{0.15 \\ $\pm$ 0.11} & \shortstack{0.12 \\ $\pm$ 0.23} & \shortstack{0.07 \\ $\pm$ 0.26} \\
     \hline
     
    \shortstack{$\theta_{P}$ ($^{\circ}$) \\ \vspace{1pt}} &
        \shortstack{-0.25 \\ $\pm$ 0.09} & \shortstack{-0.04 \\ $\pm$ 0.21} & \shortstack{0.08 \\ $\pm$ 0.14} & \shortstack{0.00 \\ $\pm$ 0.07} &
            \shortstack{-0.44 \\ $\pm$ -} & \shortstack{-0.03 \\ $\pm$ 0.23} & \shortstack{0.04 \\ $\pm$ 0.21} & \shortstack{0.03 \\ $\pm$ 0.08} & 
                \shortstack{-0.13 \\ $\pm$ 0.12} & \shortstack{-0.05 \\ $\pm$ 0.22} & \shortstack{0.07 \\ $\pm$ 0.14} & \shortstack{0.00 \\ $\pm$ 0.06} & 
                    \shortstack{0.20 \\ $\pm$ 0.15} & \shortstack{-0.03 \\ $\pm$ 0.23} & \shortstack{-0.21 \\ $\pm$ 0.11} & \shortstack{0.02 \\ $\pm$ 0.07} \\
     \hline
     
    \shortstack{$\theta_{Y}$ ($^{\circ}$) \\ \vspace{1pt}} & 
        \shortstack{0.13 \\ $\pm$ 0.18} & \shortstack{-0.06 \\ $\pm$ 0.18} & \shortstack{-0.13 \\ $\pm$ 0.18} & \shortstack{0.14 \\ $\pm$ 0.18} & 
            \shortstack{0.1 \\ $\pm$ -} & \shortstack{-0.03 \\ $\pm$ 0.20} & \shortstack{-0.03 \\ $\pm$ 0.19} & \shortstack{0.03 \\ $\pm$ 0.23} & 
                \shortstack{0.8 \\ $\pm$ 0.20} & \shortstack{-0.06 \\ $\pm$ 0.20} & \shortstack{-0.21 \\ $\pm$ 0.16} & \shortstack{0.12 \\ $\pm$ 0.21} & 
                    \shortstack{-0.11 \\ $\pm$ 0.14} & \shortstack{-0.10 \\ $\pm$ 0.22} & \shortstack{-0.14 \\ $\pm$ 0.20} & \shortstack{0.02 \\ $\pm$ 0.22} \\
     \hline
     
    \shortstack{$t_{x}$ (cm) \\ \vspace{1pt}} & 
        \shortstack{-8.0 \\ $\pm$ 4.6} & \shortstack{9.0 \\ $\pm$ 4.5} & \shortstack{0.0 \\ $\pm$ 3.6} & \shortstack{-0.1 \\ $\pm$ 2.8} & 
            \shortstack{3.7 \\ $\pm$ -} & \shortstack{9.4 \\ $\pm$ 4.4} & \shortstack{0.3 \\ $\pm$ 3.8} & \shortstack{1.5 \\ $\pm$ 3.5} & 
                \shortstack{-5.7 \\ $\pm$ 5.7} & \shortstack{9.0 \\ $\pm$ 4.9} & \shortstack{1.1 \\ $\pm$ 3.0} & \shortstack{0.1 \\ $\pm$ 3.0} & 
                    \shortstack{-2.0 \\ $\pm$ 1.8} & \shortstack{9.4 \\ $\pm$ 4.8} & \shortstack{2.2 \\ $\pm$ 2.7} & \shortstack{1.6 \\ $\pm$ 3.3} \\
     \hline
     
    \shortstack{$t_{y}$ (cm) \\ \vspace{1pt}} &
        \shortstack{-0.3 \\ $\pm$ 6.6} & \shortstack{2.1 \\ $\pm$ 4.2} & \shortstack{3.0 \\ $\pm$ 3.7} & \shortstack{5.7 \\ $\pm$ 1.5} & 
            \shortstack{3.0 \\ $\pm$ -} & \shortstack{1.3 \\ $\pm$ 4.6} & \shortstack{1.4 \\ $\pm$ 2.4} & \shortstack{5.1 \\ $\pm$ 1.7} & 
                \shortstack{1.4 \\ $\pm$ 6.8} & \shortstack{2.1 \\ $\pm$ 4.5} & \shortstack{1.8 \\ $\pm$ 2.0} & \shortstack{5.3 \\ $\pm$ 1.6} & 
                    \shortstack{4.6 \\ $\pm$ 3.3} & \shortstack{2.9 \\ $\pm$ 5.1} & \shortstack{0.0 \\ $\pm$ 3.1} & \shortstack{4.9 \\ $\pm$ 1.8} \\
     \hline
     
    \shortstack{$y_{z}$ (cm) \\ \vspace{1pt}} &
        \shortstack{-3.7 \\ $\pm$ 2.2} & \shortstack{0.2 \\ $\pm$ 4.7} & \shortstack{1.4 \\ $\pm$ 1.3} & \shortstack{-2.4 \\ $\pm$ 2.0} & 
            \shortstack{-8.2 \\ $\pm$ -} & \shortstack{0.6 \\ $\pm$ 5.0} & \shortstack{1.5 \\ $\pm$ 1.0} & \shortstack{-2.2 \\ $\pm$ 1.9} & 
                \shortstack{-1.6 \\ $\pm$ 2.6} & \shortstack{0.0 \\ $\pm$ 4.8} & \shortstack{1.5 \\ $\pm$ 1.5} & \shortstack{-1.4 \\ $\pm$ 2.0} & 
                    \shortstack{5.5 \\ $\pm$ 2.9} & \shortstack{0.6 \\ $\pm$ 5.0} & \shortstack{2.0 \\ $\pm$ 1.0} & \shortstack{-1.5 \\ $\pm$ 2.2} \\
    \hline
    \end{tabular}}
\end{center}
\caption{Rotation and translation extrinsic calibration optimization results using I2I and D2D mutual information at four different error levels. D2D optimization has significantly more "hits" and is more precise than I2I optimization. However the optimization is noticeably more sensitive to rotation error than in the rotation optimization only case. Statistics are calculated only for optimizations that hit.}
\label{table:results_rotation_and_translation}
\end{table*}

\subsection{Rotation and Translation Experiment}
In this experiment we model the error as a rotation and translation of the LiDAR. This is the general case where no constraint is made on the motion of the camera or LiDAR. Here only a weak prior on the translation component is available. We model this with two translation error ranges of $25$cm and $50$cm. 

The results of this experiment are shown in \cref{table:results_rotation_and_translation}. Clearly the 6 DoF case is more challenging than the 3 DoF case. However, the increased hit rate of D2D with respect to I2I is again clear. For example at $0.5^{\circ}$ and $50$cm error only 9.5\% and 22\% of I2I executions hit whereas for D2D 88\% and 76.5\% of executions hit for KITTI and KITTI-360 respectively.

Again the precision of D2D is higher than I2I. The mean values of the D2D translation parameters across the four different error initialization vary by less than a magnitude of $2$cm. For example, for the D2D KITTI $t_{x}$ parameter, the mean value is $9.0$, $9.4$, $9.0$ and $9.4$cm across the four error ranges. In contrast to the I2I values of $-8.0$, $3.7$, $-5.7$ and $-2.0$cm. This consistency pattern is also seen in the value of the rotation parameters.

The low hit rate of I2I makes it challenging to compare the accuracy of I2I to D2D at any error range. Furthermore, the large deviation in the mean value makes direct comparison of the mean values less meaningful. However, D2D can recover the value of the ground truth calibration to within $0.18^{\circ}$ and $9.9$cm for KITTI and $0.08^{\circ}$ and $5.4$cm for KITTI-360.

\section{Conclusion}
We have presented a robust and explainable camera-LiDAR extrinsic calibration algorithm using geometric mutual information. We introduced three contributions: (i) the use of monocular depth estimation networks as a virtual sensor for camera-LiDAR extrinsics calibration, (ii) the use of geometric mutual information for camera LiDAR extrinsic calibration, (iii) and the first extrinsic calibration experiments on the KITTI-360 fisheye data set. Our proposed algorithm requires no ground truth data, hand engineered features or offline optimization and is suitable for the continuous online extrinsic calibration (COEC) of camera-LiDAR sensors found in automated vehicles and robotics systems.

\subsection{Future Work}
Further work in this area should include:

\begin{itemize}
    \itemsep0em 
    \item Experimental results exploring the temporal consistency and data dependency
    \item Understanding the impact of occlusion and large baseline camera-LiDAR systems
    \item Exploration of optimization strategies for the 6 DoF case
    \item Research on the effect of degraded monocular depth estimation quality
    \item Geometric camera-LiDAR mutual information for continuous online intrinsic camera calibration (COIC)
\end{itemize}

\FloatBarrier
{\small
\bibliographystyle{ieee_fullname}
\bibliography{wacv2024}
}

\end{document}